\documentclass[11pt]{article}
\usepackage[dvipsnames]{xcolor}
\usepackage{acl}
\usepackage{times}
\usepackage{latexsym}
\usepackage[T1]{fontenc}
\usepackage[utf8]{inputenc}
\usepackage{microtype}
\usepackage{hyperref}
\usepackage{todonotes}
\usepackage{makecell}

\usepackage{graphicx}
\usepackage[export]{adjustbox}
\usepackage{enumitem}
\usepackage{rotating}
\usepackage{arydshln}
\usepackage{multirow}
\usepackage{multicol}
\usepackage{booktabs}
\usepackage{tabularx}

\newcommand{\perc}{\footnotesize{\%}}

\title{DyKnow: \underline{Dy}namically Verifying Time-Sensitive Factual \underline{Know}ledge \\ in LLMs}

\author{Seyed Mahed Mousavi, Simone Alghisi, Giuseppe Riccardi\\
         Signals and Interactive Systems Lab, University of Trento, Italy \\
        \texttt{ \{mahed.mousavi,s.alghisi,giuseppe.riccardi\}@unitn.it}}

\begin{document}
\maketitle
\begin{abstract} 
LLMs acquire knowledge from massive data snapshots collected at different timestamps. Their knowledge is then commonly evaluated using static benchmarks. However, factual knowledge is generally subject to time-sensitive changes, and static benchmarks cannot address those cases. We present an approach to dynamically evaluate the knowledge in LLMs and their time-sensitiveness against Wikidata, a publicly available up-to-date knowledge graph. We evaluate the time-sensitive knowledge in twenty-four private and open-source LLMs, as well as the effectiveness of four editing methods in updating the outdated facts. Our results show that 1) outdatedness is a critical problem across state-of-the-art LLMs; 2) LLMs output inconsistent answers when prompted with slight variations of the question prompt; and 3) the performance of the state-of-the-art knowledge editing algorithms is very limited, as they can not reduce the cases of outdatedness and output inconsistency.

\end{abstract}

\section{Introduction}

Large Language Models (LLMs) have been compared to traditional \textit{knowledge repositories}, such as knowledge bases, knowledge graphs, and search engines regarding their capability to retrieve factual knowledge \cite{cohen-etal-2023-crawling, sun2023head, pinter-elhadad-2023-emptying,hu2024towards}. A critical requirement for a reliable knowledge repository is to maintain the accuracy of the factual information it contains. The factual knowledge has a dynamic nature and can change significantly from what has been first inserted; and in the case of LLMs what was observed during the training stage. 

\begin{figure}[t]
    \includegraphics[width=0.92\linewidth]{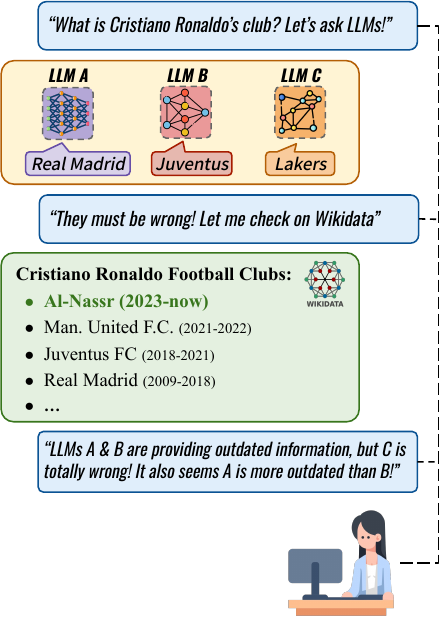}
    \caption{LLMs \textit{A}, \textit{B}, and \textit{C}  may respond with outdated (\textit{Real Madrid, Juventus}) and irrelevant (\textit{Lakers}) responses, respectively, to the user question:"\textit{What is Cristiano Ronaldo's club?}". Wikidata contains up-to-date information to assess the models' accuracy and time-sensitiveness.}
    \label{firstfigure}
\end{figure} 

LLMs are static models that are prone to generating invalid and contradicting information, and eventually getting outdated over time. They derive their knowledge from vast and often unoptimized collections of data snapshots, that are collected at different timestamps \cite{dhingra-etal-2022-time}, and typically contain substantial overlaps \cite{soldaini2024dolma}. These collections contain factual associations interspersed with inaccuracies, outdated information, and contradictions.

The maintenance of LLMs' knowledge requires a systematic approach to \textit{i}) identifying the outdated knowledge, \textit{ii}) locating it within the model parameters, and \textit{iii}) applying the necessary changes. There have been interesting studies on locating the factual associations in LLMs \cite{meng2022locating}, understanding how they are retrieved \cite{geva-etal-2023-dissecting}, and editing them \cite{li2024unveiling}. However, there are no studies on detecting outdated knowledge in LLMs (\textit{i}). While several static benchmarks have been proposed to assess the factuality of LLMs \cite{hu2024towards}, such benchmarks are not suitable for detecting outdated knowledge in LLMs. Due to the dynamic nature of knowledge, a static benchmark can quickly become outdated and lose its relevance. Moreover, it can be prone to leakage into the training data of future models (contamination). Consecutively, studies on editing techniques are mainly based on annotated edit-target datasets of synthetically generated counterfacts, leaving a gap in understanding how these methods perform with real-world data across diverse domains \cite{zhang-etal-2023-large}.

To address the issues of static benchmarks, dynamic benchmarking has been proposed where the data points are continuously updated to reflect real-time scenarios. Despite extensive research on static benchmarking, theoretical and empirical research on dynamic benchmarking is very limited \cite{shirali2023a}, making it challenging and expensive to construct a valid dynamic benchmark \cite{yin-etal-2023-alcuna}. We present an approach to dynamically benchmark the factual knowledge in LLMs using Wikidata\footnote{\href{https://github.com/sislab-unitn/DyKnow}{Link to our Repository}}.  

For each factual association in the form of \textit{(subject, property, attribute)}, the most current attribute values are obtained from the Wikidata knowledge base at the time of evaluation, in addition to the complete list of outdated values along with their validity interval (for example, in Figure \ref{firstfigure}, the validity interval of "{\fontfamily{cmtt}\selectfont{Juventus FC}}" as the correct attribute for "{\fontfamily{cmtt}\selectfont{Cristiano Ronaldo's current football club}}" is 2018-2021). The attribute value generated by the model is then validated against this comprehensive list, evaluating the accuracy and timeliness of the model responses. 

We assess the efficacy of the proposed approach by investigating the following Research Questions (RQs):

\begin{itemize} [noitemsep,topsep=1pt,parsep=1pt,partopsep=1pt]
    \item \textbf{\textit{RQ1. How reliable are state-of-the-art LLMs in responding to time-sensitive factual questions?}} We evaluate the knowledge of 24 LLMs regarding a diverse set of time-sensitive facts. We further evaluate the consistency of the model outputs across various prompts, as an indicator of input-bound uncertainty \cite{portillo-wightman-etal-2023-strength,lyu2024calibrating}. 
    \item \textbf{\textit{RQ2. Can we estimate the temporal interval of the data used to (pre-)train the LLMs?}} We analyze the outputs of each model based on their validity intervals and approximate the temporal interval of the (pre-)training data. We compare our estimations with the reports from models that have disclosed details of their (pre-)training data.
    \item \textbf{\textit{RQ3. Can knowledge editing methods improve the accuracy and consistency of LLMs regarding real-world time-sensitive facts?}} We select four outdated LLMs and apply four editing algorithms to update their outdated knowledge regarding the real world. We evaluate the effectiveness and scalability of the editing algorithms in updating LLMs regarding real-world facts. 
\end{itemize}

\section{Literature Review}

\textbf{LLMs as Knowledge Repositories} \citet{pinter-elhadad-2023-emptying} noted that current LLMs fall short as knowledge repositories due to issues with \textit{editing}, \textit{logical consistency}, \textit{reasoning}, and \textit{interoperability}. They identified problems with existing knowledge editing techniques, such as catastrophic forgetting \cite{ratcliff1990connectionist}, limitations on the number of edits \cite{mitchell2021fast}, ripple effect failures \cite{cohen2023evaluating}, and lack of robustness \cite{brown2023robustness, hase2023does}. \citet{mazzia2023survey} summarized model editing research across computer vision and NLP fields. \citet{zhang-etal-2023-large} studied methods for aligning LLMs with real-world knowledge, pointing out issues such as unrealistic evaluation settings, synthetic datasets, insufficient quantitative analysis, and lack of studies on detecting outdated knowledge in LLMs.

\textbf{Knowledge Benchmarks} Studies on temporal reasoning in LLMs evaluate the knowledge of the model regarding a specific time in the past via an explicit time-specifier \cite{chen2021dataset, gupta-etal-2023-temptabqa}, or in more challenging settings multiple temporal factors \cite{wei-etal-2023-menatqa}. \citet{yu2023kola} proposed an evaluation setup assessing models on memorization, understanding, application, and creation of knowledge. \citet{yin-etal-2023-alcuna} discussed challenges in building dynamic factual benchmarks and suggested generating artificial new knowledge by randomly altering entities/relations within the same ontological class. While the mentioned studies presented static benchmarks, \citet{kasai2022realtime} introduced RealTime QA, a benchmark with 30 weekly questions and answers for LLM evaluation. Meanwhile, \citet{jang-etal-2022-temporalwiki} presented an approach to track changes in knowledge by comparing consecutive snapshots of Wikipedia and re-training the models on the identified differences. 


\section{DyKnow \includegraphics[height=2.75ex,valign=m]{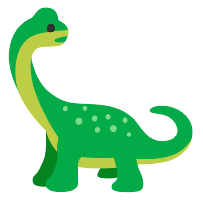}: Dynamic Knowledge Validation}

The benchmark for evaluating time-sensitive knowledge in LLMs must be model-agnostic and long-lasting since it must not become outdated as the models. A static benchmark lacks these characteristics, cannot capture the changing world, and can lead to data contamination. 

We present a cost-effective approach to dynamically benchmark the factual knowledge in LLMs using Wikidata. Wikidata is a multilingual knowledge graph that is continuously and collaboratively edited to maintain up-to-date information \cite{vrandevcic2014wikidata}. Factual Knowledge in Wikidata is presented by \textit{properties} that connect \textit{subject} nodes to \textit{attribute} values. For example, "Cristiano Ronaldo's current football club" factual knowledge is presented by the property {\fontfamily{cmtt}\selectfont{"member of sports team"}} that connects the subject {\fontfamily{cmtt}\selectfont{"Cristiano Ronaldo"}} to the current attribute value, at the time of paper "{\fontfamily{cmtt}\selectfont{Al-Nassr}}". Furthermore, the \textit{attribute} values in Wikidata are accompanied by \textit{qualifiers}, which provide additional context and specificity regarding the attribute values such as geographical locations, measurement units, as well as start and end dates for attribute values that have a temporal validity interval. For instance, the attribute value "{\fontfamily{cmtt}\selectfont{Al-Nassr}}" is accompanied by start and end date quantifiers  {\fontfamily{cmtt} \selectfont{"2023-Now"}}. Besides the current attribute value for each factual triplet, Wikidata maintains all the previously correct attribute values in addition to their corresponding start and end date quantifiers, indicating the corresponding temporal validity intervals. Therefore, the complete list of attributes for "Cristiano Ronaldo's current football club" factual knowledge consists of [{\fontfamily{cmtt}\selectfont{Al-Nassr \textsubscript{2023-Now}, Manchester United F.C. \textsubscript{2021-2022}, Juventus FC \textsubscript{2018-2021}, Real Madrid \textsubscript{2009-2018},...}}] (Figure \ref{firstfigure}). 

Instead of relying on static ground truth values, we evaluate the models' outputs with the list of attribute values retrieved dynamically from the Wikidata knowledge base at the time of evaluation. We assess the knowledge of the model regarding each fact as:  
\begin{itemize}[noitemsep,topsep=1pt,parsep=1pt,partopsep=1pt]
    \item \textbf{Correct} when the model outputs the most up-to-date value from the list; we further categorize the \textit{Incorrect} outputs of the model as  
    \item \textbf{Outdated} when the model outputs a value that is not correct anymore and now is outdated; and 
    \item \textbf{Irrelevant} when the model output is not present in the Wikidata list (e.g. due to hallucination or contradicting/false information in the training data)
\end{itemize} 

Furthermore, by analyzing the correct and outdated outputs of each model according to their validity interval, we can approximate the temporal interval of the data used for (pre-)training the models. For instance,  if a model provides outdated responses to time-sensitive questions, with the oldest responses dating back to 2016 and the most recent ones correct until 2019, we can infer that the model was likely trained on data collected up to 2020, encompassing documents from 2016 to 2019.

\section{Validating DyKnow \includegraphics[height=2.75ex,valign=m]{emnlpImages/sauropod.pdf}}

To assess the efficacy of the proposed approach, we evaluate 24 LLMs on 130 time-sensitive facts including countries' politicians, athletes' clubs, and organizations' roles. This allow us to introduce diversity in the dataset by having human subjects, organization subjects, and country subjects with a diverse set of properties to query the models.

\textbf{Time-Sensitive Facts} We aim to select subject entities that are most likely to be frequently present in the training data of most LLMs. This choice is motivated by studies showing the performance of LLMs regarding factual information about an entity depends on its frequency in the training data \cite{pinter-elhadad-2023-emptying, mallen-etal-2023-trust}. We select the top 50 countries by Gross Domestic Product (GDP) in 2023, the top 30 athletes of 2023 (10 soccer players, 10 basketball players, and 10 Formula 1 drivers), and 25 public and private organizations (the top 20 companies by revenue and the top 5 organizations by influence). For each country, we query the models about the "\textit{head of state}" (e.g., president, king) and the "\textit{head of government}" (e.g., prime minister, premier). For each athlete, we query about their \textit{sports team}, and for each organization, we ask about the corresponding \textit{directorial role} (e.g., CEO, chairperson). After manually removing the subjects with missing property/attributes in Wikidata, the final list of time-sensitive facts to evaluate the LLMs consists of 78 facts about 47 countries, 28 facts about 28 athletes, and 24 facts about 23 organizations. The complete list of subject entities and properties as time-sensitive facts used in benchmarking the LLMs is presented in § Table \ref{tab:subs}.

\textbf{LLMs} We evaluate the following 24 LLMs: GPT-2 XL \cite{radford2019language}, GPT-3\footnote{\href{https://platform.openai.com/docs/models/gpt-base}{davinci-002}} \cite{brown2020language}, T5 (3B) \cite{raffel2020exploring}, GPT-J (6B) \cite{gpt-j}, ChatGPT (GPT-3.5)\footnote{\href{https://platform.openai.com/docs/models/gpt-3-5}{gpt-3.5-turbo-1106}}, Bloom (7B) \cite{workshop2022bloom}, Flan-T5 XL \cite{chung2022scaling}, GPT-4\footnote{\href{https://platform.openai.com/docs/models/gpt-4-and-gpt-4-turbo}{gpt-4-1106-preview}}, Llama-2 (7B) \& Llama-2 Chat (7B) \cite{touvron2023llama}, Falcon (7B) \& Falcon Instruct (7B) \cite{almazrouei2023falcon}, Vicuna v1.5 (7B) \cite{vicuna2023}, Mistral v0.1 (7B) \& Mistral Instruct v0.1 (7B) \cite{jiang2023mistral}, Mixtral 8x7B v0.1 \& Mixtral 8x7B Instruct v0.1 \cite{jiang2024mixtral}, OLMo (1B \& 7B) \cite{groeneveld2024olmo}, Llama-3 (8B) and Llama-3 Instruct (8B)\footnote{\href{https://ai.meta.com/blog/meta-llama-3/}{LLaMA-3}}, OpenELM (270M \& 1.1B \& 3B)~\cite{mehta2024openelm}. 

\section*{RQ1: LLMs' Time-Sensitive Knowledge}

\subsection*{A. Knowledge Evaluation}
Using DyKnow, we evaluate 24 LLMs regarding 130 time-sensitive facts about frequent subject entities in different categories (human subjects, organization subjects, and country subjects). For each time-sensitive fact, the outputs of the models are validated against a list of attribute values dynamically retrieved from Wikidata, classifying the outputs as \textbf{C}orrect, \textbf{O}utdated, and \textbf{I}rrelevant.

\begin{table}[t!]
\centering
    \begin{adjustbox}{width=1.05\linewidth,center=\linewidth}
        \begin{tabularx}{1.05\linewidth}{lrrr}
            \toprule
            \textbf{(Year) Model} & \textbf{C}{\scriptsize orrect}& \textbf{O}{\scriptsize utdated} &\textbf{I}{\scriptsize rrelevant} \\
            \midrule
            \noalign{\smallskip} 
            {\small(2019)} GPT-2 & 26\perc & 42\perc & 32\perc\\
            {\small(2020)} GPT-3 & 42\perc & 47\perc & 12\perc\\
            {\small(2020)} T5 & 11\perc & 21\perc & 68\perc\\
            {\small(2021)} GPT-J & 41\perc & 46\perc & 13\perc\\
            {\small(2022)} Bloom & 35\perc & 49\perc & 16\perc\\
            {\small(2022)} Flan-T5 & 18\perc & 39\perc & 43\perc\\
            {\small(2023)} Llama-2 & 51\perc & 42\perc & 7\perc\\
            {\small(2023)} Falcon & 42\perc & 47\perc & 11\perc\\
            {\small(2023)} Mistral & 53\perc & 39\perc & 8\perc\\
            {\small(2023)} Mixtral & 48\perc & 42\perc & 10\perc\\
            {\small(2024)} OLMo \small{1B} & 37\perc & 40\perc & 23\perc\\
            {\small(2024)} OLMo \small{7B} & 35\perc & 36\perc & 29\perc\\
            {\small(2024)} Llama-3 & 57\perc & 36\perc & 7\perc\\
            {\small(2024)} OpenELM \small{270M} & 12\perc & 28\perc & 61\perc\\
            {\small(2024)} OpenELM \small{1.1B} & 35\perc & 47\perc & 18\perc\\
            {\small(2024)} OpenELM \small{3B} & 42\perc & 42\perc & 16\perc\\
            \noalign{\smallskip} 
            \cdashline{1-4}
            \noalign{\smallskip} 
            {\small(2022)} ChatGPT & 57\perc & 35\perc & 8\perc\\
            {\small(2023)} GPT-4 & 80\perc & 13\perc & 7\perc\\
            {\small(2023)} Llama-2$_{C.}$ & 51\perc & 37\perc & 12\perc\\
            {\small(2023)} Falcon$_{I.}$ & 44\perc & 41\perc & 15\perc\\
            {\small(2023)} Vicuna & 52\perc & 33\perc & 15\perc\\
            {\small(2023)} Mistral$_{I.}$ & 52\perc & 32\perc & 16\perc\\
            {\small(2023)} Mixtral$_{I.}$ & 62\perc & 29\perc & 9\perc\\
            {\small(2024)} Llama-3$_{I.}$ & 76\perc & 14\perc & 10\perc\\
            \bottomrule
        \end{tabularx} 
    \end{adjustbox}
    \caption{Benchmarking 24 LLMs with time-sensitive knowledge via \textit{Upper Bound}. The table presents the percentage of \textbf{C}{orrect} answers that are valid and up-to-date, \textbf{O}utdated answers that are not valid anymore, and \textbf{I}rrelevant outputs. Models below the dashed line were prompted with an additional prefix "Answer with the name only". Subscripts ${I.}$ and ${C.}$ stand for \textit{Instruct} and \textit{Chat}, respectively.}
\label{table:full_currency}
\end{table}

\textbf{Prompting Strategy} We develop a prompt template for each time-sensitive fact and subject group, including placeholders for subject names and, for countries, official titles. Using GPT-4, we generate four rephrased versions of each prompt as slightly perturbed lexicalizations and ask three human judges (researchers in our group) to review and validate the generated prompts. After collecting feedback and manual controls, three question prompts are selected for each fact. We then queried the models for each time-sensitive fact using the selected three prompts. In contrast to studies on the temporal reasoning of LLMs \cite{chen2021dataset, wei-etal-2023-menatqa, gupta-etal-2023-temptabqa}, our questions are framed in the present tense, omit explicit time specifiers, and seek the \textit{currently correct} answer. § Table \ref{tab:subs} presents the prompt templates used to query the models for time-sensitive facts for each subject category.

\textbf{Upper Bound} We validate the generated outputs using an "Upper Bound" approach. If the model provides the correct (up-to-date) answer to at least one of the three prompts, we consider it a success, indicating that the information in the model regarding that specific fact is current. If the model fails to give a correct answer but provides an outdated response to at least one of the prompts, we classify the information in the model as outdated. Irrelevant outputs may occur due to several reasons: a) the model may not have learned the specific time-sensitive fact during (pre-)training or fine-tuning, b) hallucinations or conflicting/false information in the training data, or c) the information may not be retrievable using our prompts.

\textbf{Results} Table \ref{table:full_currency} shows the results of this evaluation on 24 LLMs\footnote{The models are evaluated with the answer sets retrieved on 18 December 2023.}. The results highlight concerning issues regarding the currency of the models' knowledge about frequent subject entities. Even the best-performing models exhibit non-negligible percentages of outdated and irrelevant answers, which can be problematic in real-world applications where up-to-date and accurate information is crucial. GPT-4 (2023) demonstrates a high rate of correct responses, while 20\% of its outputs are either outdated or irrelevant. Similarly, more recent models such as Llama-3 (2024), OLMo (2024), and OpenELM (2024) output incorrect (outdated and irrelevant) responses to more than 40\% of the questions. As expected, older models like GPT-2 (2019) and GPT-3 (2020) demonstrate lower levels of up-to-dateness. These statistics imply that a significant portion of the models' outputs are either outdated or irrelevant, potentially leading to misinformation if relied upon.


\subsection*{B. Output Consistency}

The consistency of model outputs across various prompts, known as \textit{prompt agreement}, has been examined in the literature as an indicator of input-bound uncertainty \cite{portillo-wightman-etal-2023-strength,lyu2024calibrating}. These studies are based on the premise that higher consistency across different prompts signals lower uncertainty in the model prediction. By querying the LLMs for each time-sensitive fact using the selected three prompts (§ Table \ref{tab:subs}), we observe that they often generate inconsistent answers to slightly modified versions of the same prompt. 


\begin{figure}[t]
    \includegraphics[trim={0.7cm 0 0 0},clip,width=\linewidth]{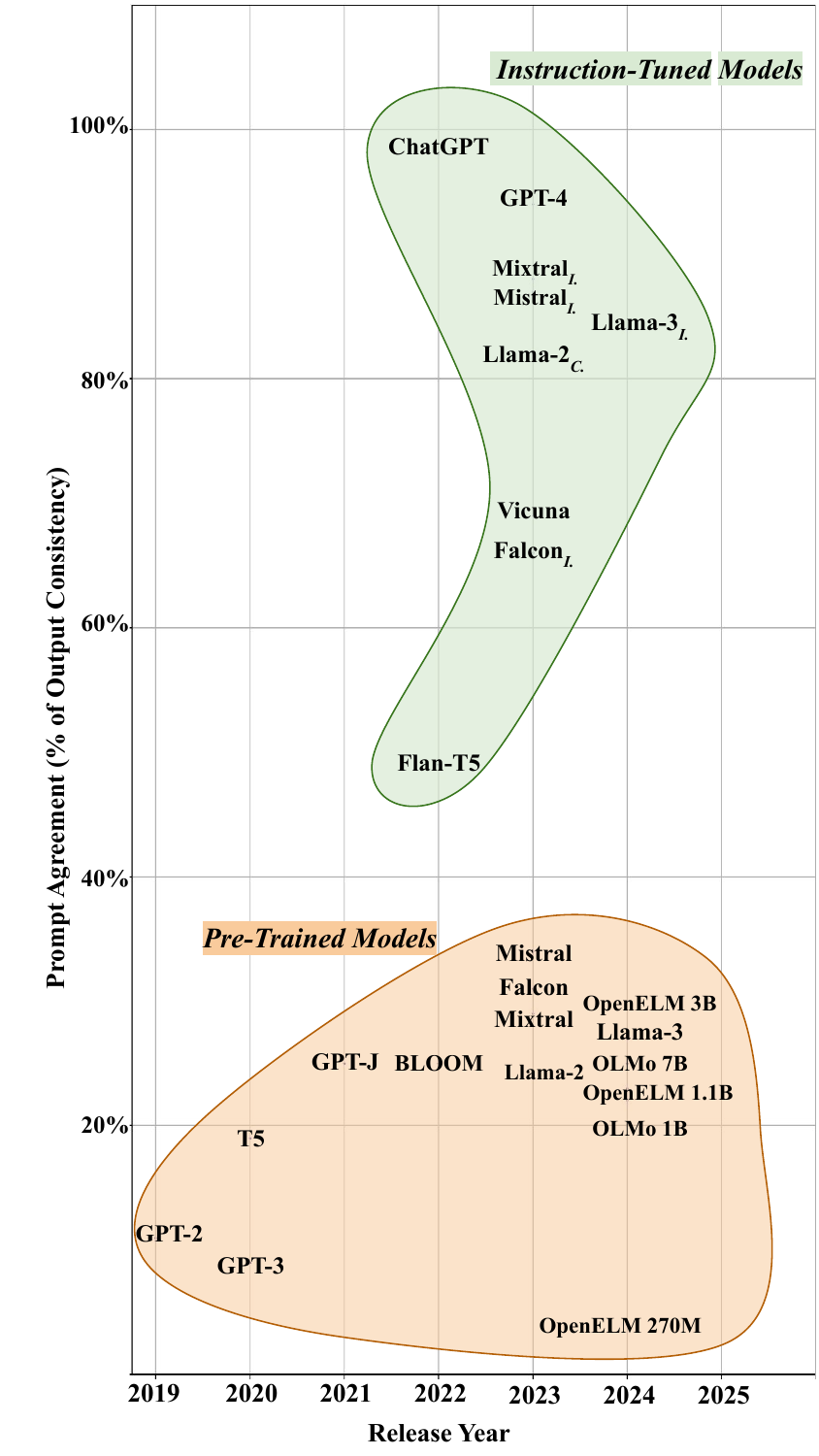}
    \caption{The level of prompt agreement for each model across three prompts for each time-sensitive question. Subscripts ${I.}$ and ${C.}$ stand for \textit{Instruct} and \textit{Chat}, respectively. Instruction-tuned models demonstrate a comparatively higher prompt agreement.}
    \label{consistencyfig}
\end{figure} 

\textbf{Results} Figure \ref{consistencyfig} presents the prompt agreement level, i.e. the consistency of outputs across different prompts for all models (the agreement percentage for each model in presented in § Table \ref{table:promptagreement}). 
The results show that prompt agreement varies significantly across different models, with most LLMs demonstrating low levels of prompt agreement, indicating that they produce varying responses to slightly altered versions of the same question. There is a trend of improvement in prompt agreement over time, with more recent models showing higher consistency in their responses. Furthermore, instruction-tuned models demonstrate a comparatively higher prompt agreement. ChatGPT (2022) and GPT-4 (2023) exhibit the highest prompt agreement. Other high performers include Mistral$_{I.}$ (2023), Mixtral$_{I.}$ (2023), and Llama-3$_{I.}$ (2024). In contrast, OpenELM 270M (2024) has the lowest agreement among the models. These results highlight the high sensitivity in the auto-regressive generation process can lead to different, incorrect, or irrelevant outputs.

\subsection*{RQ2: LLMs' Data Interval Approximation}

\begin{figure*}[t]
    \centering
    \includegraphics[width=\textwidth]{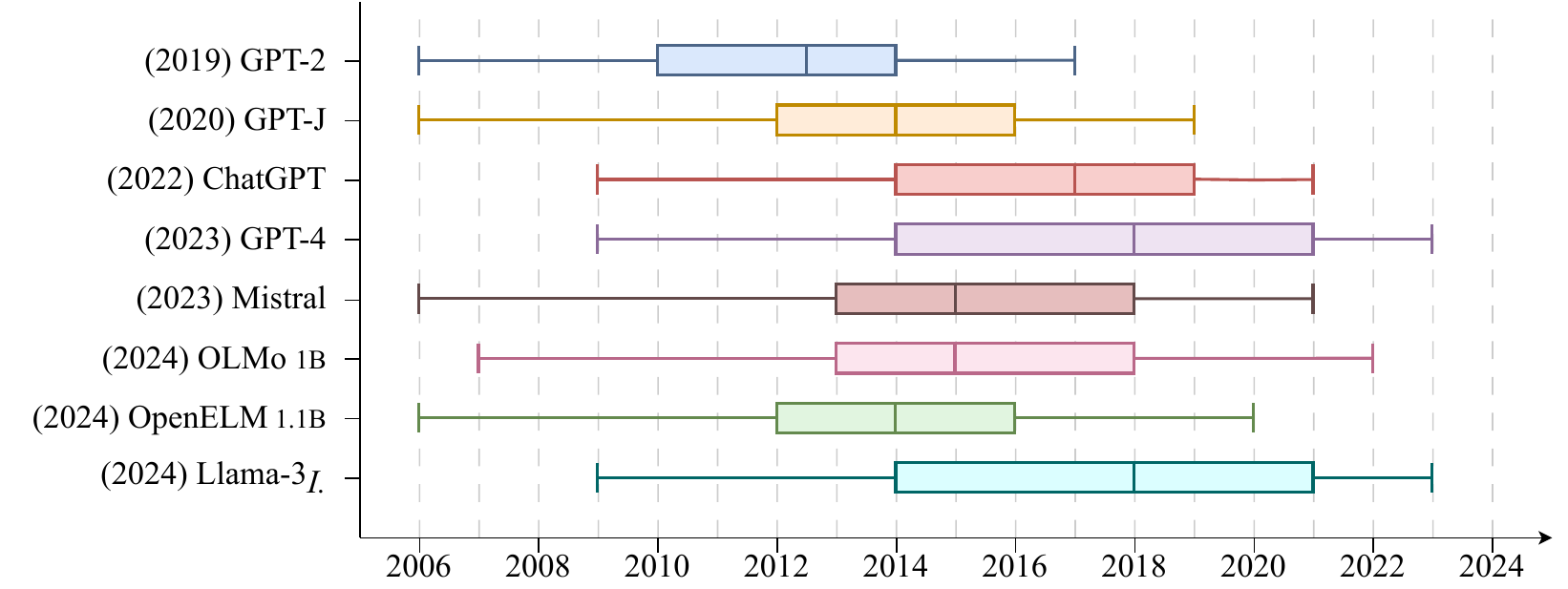}
    \caption{Approximating the temporal interval of the data used for (pre-)training LLMs following our evaluation regarding time-sensitive knowledge. The y-axis presents the evaluated LLMs with their release year in parentheses. The box plots present the distribution of the generated responses for each LLM according to their validity interval. For instance, the responses of OpenELM {\small{1.1B}} range from 2006 to 2020, with a concentrated period between 2012 and 2016, suggesting that the mode is trained on comparatively older datasets.}
    \label{fig:observedyears}
\end{figure*}

Each attribute value in Wikidata is accompanied by start and end date quantifiers, indicating the corresponding temporal validity intervals. We analyze the correct and outdated outputs of each model according to their temporal validity intervals and approximate the temporal interval of the data used for (pre-)training the models. 

\textbf{Results} The results of this analysis for GPT-\{2,J,4\}, ChatGPT, Mistral, OLMo {\small{1B}}, OpenELM {\small{1.1B}}, and Llama-3$_{I.}$ are presented in Figure \ref{fig:observedyears} (The results for the remaining LLMs are presented in § Figures \ref{fig:observedyears2} and  \ref{fig:observedyears3}). Regarding the GPT model family, we approximate that older models such as GPT-\{2,J\} are trained on older datasets as a considerable portion of their responses date back to before 2009, contributing to their outdatedness compared to relatively new models, i.e. ChatGPT, GPT-4. There is a trend of improvement in the GPT family over time, as each model demonstrates a more recent median and maximum date compared to preceding models. While the maximum data value for ChatGPT is 2021, GPT-4 has generated responses with information from 2022 and 2023. This finding aligns with the OpenAI API report, which states that the training data for ChatGPT includes information "up to September 2021", while the training data for GPT-4 includes information "up to April 2023" \footnote{\href{https://platform.openai.com/docs/models/gpt-4-and-gpt-4-turbo}{OpenAI API Link}}. Regarding recently released models, OLMo {\small{1B}} generates a broad range of responses from 2006 to 2022 with the central part of the data from 2013 to 2018. This finding suggests that the model is (pre-)trained on a wide span of data and is in line with OLMo {\small{1B}} data sheet paper \cite{soldaini2024dolma}. Llama-3$_{I.}$ demonstrates the same temporal distribution as GPT-4. Instead, the responses of OpenELM {\small{1.1B}} range from 2006 to 2020, with a concentrated period between 2012 and 2016, suggesting that the model is trained on comparatively older datasets. In general, this analysis indicates that more recent models tend to include data from the last few years, leading to potentially more correct outputs. However, the presence of outdated responses in models highlights the importance of regular updates to maintain the currency and accuracy of the (pre-)training data and the models.

\subsection*{RQ3: Updating LLMs' Knowledge}

Studies on editing techniques primarily rely on annotated edit-target datasets of synthetically generated counterfacts, leaving a gap in understanding their performance with real-world data across diverse domains \cite{zhang-etal-2023-large}. To bridge this gap, we select four outdated LLMs and evaluate the efficacy of four editing algorithms to update their outdated knowledge on real-world data. Regarding the LLMs, we have selected GPT-\{2,J\} due to generating a high percentage of outdated responses among models in Table \ref{table:full_currency}; and Llama-2$_{C.}$ and Mistral$_{I.}$ since, despite being relatively new models, provide outdated information to around 30\% of the questions. 

\begin{table*}[h!]
\centering
\small
    \begin{tabular}{lccccccc}
        \toprule
        \multirow{4}{*}{\textbf{Model}} & \multirow{4}{*}{\textbf{\makecell{\#Outdated\\Facts}}} & & \multicolumn{5}{c}{\textbf{Knowledge Editing}} \\
        \cmidrule(rrrr){4-8}
        & & & \multicolumn{2}{c}{\textbf{Modifying Parameters}} & & \multicolumn{2}{c}{\textbf{Preserving Parameters}} \\
        \cmidrule(rr){4-5} \cmidrule(rr){7-8}
        \noalign{}
        & & & {ROME} & {MEMIT} & & {SERAC} & {IKE} \\
        \midrule
        ({\small{2019}}) GPT-2 & 54 & & 17\perc & 33\perc & & 4\perc & 49\perc \\

        ({\small{2021}}) GPT-J & 60 & & 11\perc & 83\perc & & 0\perc & 97\perc\textsuperscript{\textdaggerdbl} \\

        ({\small{2023}}) Llama-2$_{C.}$ & 48 & & 4\perc & 77\perc & & 36\perc & 18\perc \\

        ({\small{2023}}) Mistral$_{I.}$ & 41 & & 0\perc & 0\perc & & --- & 92\perc\textsuperscript{\textdaggerdbl} \\
        \bottomrule
        \end{tabular} 
    \caption{Performance of different knowledge editing methods for updating the outdated facts in LLMs, by the \textit{harmonic mean} of efficacy success and paraphrase success \cite{meng2022locating, meng2022mass}. \textbf{\textdaggerdbl} indicates successful alignment of more than 85\% outdated knowledge.}
\label{table:edits_final}
\end{table*}

\textbf{Methods} Regarding algorithms that modify LLM parameters to incorporate edited knowledge, we evaluate two methods. First, \textbf{ROME} \cite{meng2022locating} locates relevant parameters in the feed-forward layers and inserts new key-value associations as a least squares problem with a linear equality constraint. Second, \textbf{MEMIT} \cite{meng2022mass} extends ROME to apply multiple edits simultaneously by operating on several layers in a single intervention. Regarding editing methods that preserve the original LLM parameters, we evaluate two approaches. First, \textbf{SERAC} \cite{pmlr-v162-mitchell22a} uses external memory to store new facts and a classifier to match question prompts with these stored facts. Depending on whether a match is found, the classifier decides whether to condition the generation of the model on the retrieved fact or not. Second, \textbf{IKE} \cite{Zheng2023CanWE} utilizes in-context learning by constructing a prompt with the question, the corresponding up-to-date fact, and a context segment consisting of examples for answering the question. The constructed prompt is then presented to the model to generate an answer. Note that this method is not entirely realistic, as it requires relevant and up-to-date facts to be provided for each question. Further implementation details of the evaluated techniques are presented in § \ref{impdet}. 

\textbf{Harmonic Mean} We evaluate the methods using the harmonic mean of efficacy success and paraphrase success \cite{meng2022locating, meng2022mass}. Efficacy success measures the proportion of correctly edited responses to the original question prompts (RQ1 A.), while paraphrase success assesses the model's performance on paraphrased versions of the prompts, serving as a generalization metric (RQ1 B.).

\begin{figure}[t!]
    \centering
    \includegraphics[width=0.85\linewidth]{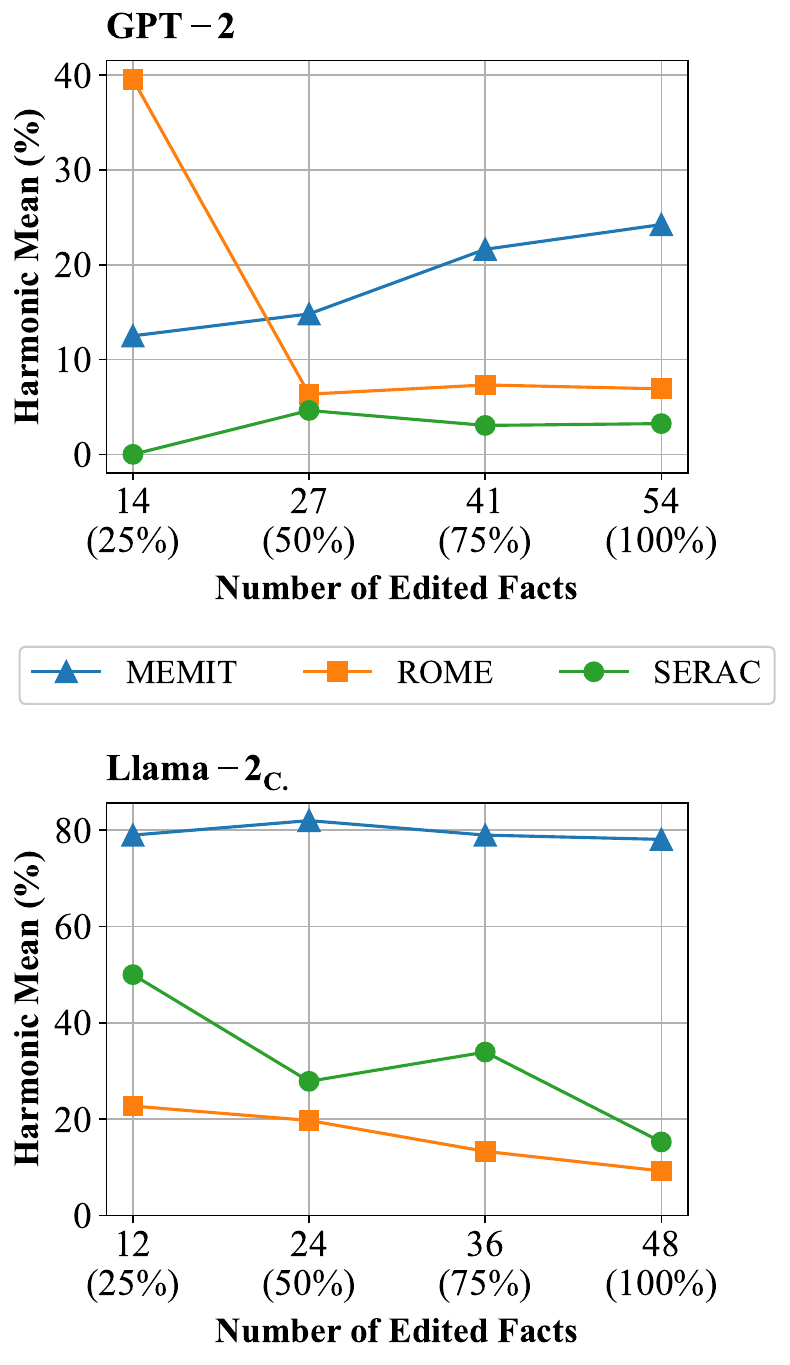}
    \caption{The scalability of editing algorithms for "\textit{updating}" the outdated facts in GPT-2 and Llama-2$_{C.}$. The x-axis and y-axis represent the number of edits (in parenthesis the percentage of the total edits) and the harmonic mean of the models, respectively.}
    \label{fig:scalability}
\end{figure}

\textbf{Results} The results, presented in Table \ref{table:edits_final}, indicate that the performance of editing methods is model-dependent (§ Table \ref{table:edits_ps} reports the performance of the methods by paraphrasing success). Among the edited LLMs, GPT-J is a better candidate for updating as it achieves a high success rate by two methods, MEMIT and IKE. Among the methods that modify the LLM parameters, ROME demonstrates an overall poor performance and MEMIT significantly outperforms ROME in most cases, especially for GPT-J and Llama-2$_{C.}$. However, none of these approaches apply to Mistral$_{I.}$ as it fails to output any meaningful sequence and generates only special tokens after parameter modification. Regarding editing methods that preserve the LLM parameters, SERAC fails to achieve high performance with updating at best less than 40\% of the outdated information in Llama-2$_{C.}$. Meanwhile, IKE achieves a high performance on GPT-J and Mistral$_{I.}$, achieving over 90\% success rates, indicating the effectiveness of in-context learning for these two models.  In general, MEMIT and IKE are the standout methods for modifying and preserving parameters, respectively. MEMIT excels with GPT-J and Llama-2$_{C.}$, while IKE shows high success rates with GPT-J and Mistral$_{I.}$. 

\textbf{Scalability Studies} We investigate the scalability of ROME, MEMIT, and SERAC in performing different subsets of edits on real-world facts in GPT-2 and Llama-2$_{C.}$. The results, shown in Figure \ref{fig:scalability}, indicate that ROME exhibits a significant decline in performance for both models as the number of edited facts increases. In contrast, MEMIT demonstrates a more stable performance, with gradual improvement on GPT-2 as the number of edits increases. SERAC, meanwhile, maintains consistently low performance on GPT-2 and shows a decline on Llama-2$_{C.}$ as the number of edits rises. Overall, ROME and SERAC exhibit poor scalability and effectiveness in handling multiple edits, while MEMIT presents stable performance across increasing numbers of edits on both models. 

\section{Discussion}

In this section, we discuss our findings in relation to the introduced research, as well as avenues for future work.

\textbf{\textit{RQ1. How reliable are state-of-the-art LLMs in responding to time-sensitive factual questions?}} While recent models like GPT-4 and Llama-3$_{I.}$ show better performance than other models, the persistent presence of outdated and incorrect information across all models suggests that current LLMs are still far from reliable knowledge sources. Furthermore, the high sensitivity of the auto-regressive generation process to slight variations in question lexicalization can lead to contradicting and sometimes incorrect or irrelevant outputs. This unreliability underscores the critical need for further refinement in training methodologies and updating mechanisms to consistently ensure these models provide accurate information at any time.

\textbf{\textit{RQ2. Can we estimate the temporal interval of the data used to (pre-)train the LLMs?}} Our approximations align with the models' reports that disclose the data used during (pre-)training. This analysis indicates that comparatively recent models tend to include data from the last few years, leading to potentially more correct outputs. However, the presence of outdated facts in all models, and thus in the (pre-)training data,  highlights the need for regular updates to maintain the currency and accuracy of the (pre-)training data and the models.

\textbf{\textit{RQ3. Can knowledge editing methods improve the accuracy and consistency of LLMs regarding real-world time-sensitive facts?}} Despite satisfactory performance on synthetic target datasets in the literature, knowledge editing methods show limitations in updating LLMs regarding real-world knowledge or improving their consistency. The model-dependent performance of the methods highlights the importance of selecting the appropriate editing technique based on the specific model in use. Furthermore, editing the knowledge in a repository requires three types of operations \cite{DIGNUM1992293}:  a) \textit{updating} an existing value attribute with a new value; b) \textit{deleting} a property/attribute thoroughly; and c) \textit{adding} a completely new property/attribute. However, studies on editing the knowledge in LLMs \cite{yao2023editing, mazzia2023survey,zhang-etal-2023-large} mostly focus on \textit{updating} operation of an existing knowledge only. This underscores the necessity for tailored approaches when editing LLMs with new knowledge to ensure accuracy and reliability. 

\section{Conclusion}

We have investigated the process of keeping LLMs' knowledge up-to-date and presented an approach to dynamically benchmarking this knowledge via Wikidata. Dynamic benchmarks are a promising solution to address the known limitations of static benchmarks, such as outdatedness and data contamination. 

Our results indicate that LLMs differ from traditional knowledge repositories, making it important to investigate what types of knowledge these models can reliably manage and what types of querying and alignment operations they support. We encourage further community engagement to expand DyKnow into a current and active benchmark.

\section*{Acknowledgement}
We acknowledge the support of the MUR PNRR
project FAIR - Future AI Research (PE00000013)
funded by the NextGenerationEU.

\section*{Limitations}
The benchmark is designed based on the Wikidata knowledge base. Other sources can be included to enrich the diversity of the domains and facts in the benchmark. The performance of the editing methods on the ripple effect of edited time-sensitive facts is not evaluated in this work. The evaluated editing methods are focused on updating the LLMs and do not consider the other operations of removing the knowledge from the model or adding knowledge to the LLM. Lastly, the evaluations of editing methods are limited due to a lack of computation resources, as we could not experiment with larger open-source models.

\bibliography{custom}
\bibliographystyle{acl_natbib}

\onecolumn

\appendix

\section{Appendix}

\begin{table*}[h!]
    \centering
    \includegraphics[width=0.97\textwidth]{emnlpImages/tableofsubjects.pdf}
    \caption{The list of subject entities and properties as time-sensitive facts used in benchmarking the LLMs. We used three prompt templates for each category. }
    \label{tab:subs}
\end{table*}

\newpage

\begin{table}[h!]
\centering
    \begin{tabularx}{0.52\textwidth}{lrrr}
        \toprule
        \textbf{(Year) model} & \textbf{C}\scriptsize{orrect}& \textbf{O}\scriptsize{utdated} &\textbf{I}\scriptsize{rrelevant} \\
        \midrule
         \noalign{\smallskip} 
        {\small(2019)} GPT-2 & 15\% & 24\% & 61\%\\
        {\small(2020)} GPT-3 & 22\% & 32\% & 46\%\\
        {\small(2020)} T5 & 5\% & 12\% & 83\%\\
        {\small(2021)} GPT-J & 29\% & 35\% & 36\%\\
        {\small(2022)} Bloom & 24\% & 36\% & 40\%\\
        {\small(2022)} Flan-T5 & 13\% & 32\% & 55\%\\
        {\small(2023)} Llama-2 & 35\% & 32\% & 33\%\\
        {\small(2023)} Falcon & 31\% & 35\% & 34\%\\
        {\small(2023)} Mistral & 38\% & 33\% & 29\%\\
        {\small(2023)} Mixtral & 36\% & 33\% & 31\%\\
        {\small(2024)} OLMo \small{1B} & 23\% & 27\% & 50\%\\
        {\small(2024)} OLMo \small{7B} & 35\% & 29\% & 36\%\\
        {\small(2024)} Llama-3 & 37\% & 34\% & 29\%\\
        {\small(2024)} OpenELM \small{270M} & 5\% & 13\% & 82\%\\
        {\small(2024)} OpenELM \small{1.1B} & 24\% & 33\% & 43\%\\
        {\small(2024)} OpenELM \small{3B} & 31\% & 31\% & 38\%\\
        \noalign{\smallskip} 
        \cdashline{1-4}
        \noalign{\smallskip} 
        {\small(2022)} ChatGPT & 56\% & 35\% & 9\%\\
        {\small(2023)} GPT-4 & 77\% & 15\% & 8\%\\
        {\small(2023)} Llama-2$_{C.}$ & 47\% & 35\% & 18\%\\
        {\small(2023)} Falcon$_{I.}$ & 38\% & 40\% & 22\%\\
        {\small(2023)} Vicuna & 44\% & 32\% & 24\%\\
        {\small(2023)} Mistral$_{I.}$ & 51\% & 29\% & 20\%\\
        {\small(2023)} Mixtral$_{I.}$ & 59\% & 31\% & 10\%\\
        {\small(2024)} Llama-3$_{I.}$ & 69\% & 17\% & 14\%\\
        \bottomrule
    \end{tabularx}
    \caption{Benchmarking 24 LLMs with time-sensitive knowledge. Differently from Table \ref{table:full_currency}, the scores are computed by averaging the model performance across the three prompts. The table presents the percentage of \textbf{C}{orrect} answers that are valid and up-to-date, \textbf{O}utdated answers that are not valid anymore, and \textbf{I}{rrelevant} outputs. Models below the dashed line were prompted with an additional prefix "Answer with the name only".}
    \label{table:full_avg_currency}
\end{table}

\begin{table}[t!]
\centering
    \begin{tabularx}{0.425\linewidth}{lr}
    \toprule
    \textbf{(Year) Model} & \textbf{\makecell{Prompt\\ Agreement (\%)}}\\
    \midrule
    {\small(2019)} GPT-2  &11\perc\\
    {\small(2020)} GPT-3  &9\perc\\
    {\small(2020)} T5  &19\perc\\
    {\small(2021)} GPT-J  &25\perc\\
    {\small(2022)} Bloom   &25\perc\\
    {\small(2022)} Flan-T5  &49\perc\\
    {\small(2023)} Llama-2   &24\perc\\
    {\small(2023)} Falcon   &31\perc\\
    {\small(2023)} Mistral   &34\perc\\
    {\small(2023)} Mixtral   &29\perc\\
    {\small(2024)} OLMo \small{1B}   &20\perc\\
    {\small(2024)} OLMo \small{7B}   &23\perc\\
    {\small(2024)} Llama-3   &25\perc\\
    {\small(2024)} OpenELM \small{270M}   &4\perc\\
    {\small(2024)} OpenELM \small{1.1B}   &22\perc\\
    {\small(2024)} OpenELM \small{3B}   &27\perc\\
    \noalign{\smallskip} 
    \cdashline{1-2}
    \noalign{\smallskip} 
    {\small(2022)} ChatGPT  &98\perc\\
    {\small(2023)} GPT-4&94\perc\\
    {\small(2023)} Llama-2$_{C.}$ &82\perc \\
    {\small(2023)} Falcon$_{I.}$  &66\perc\\
    {\small(2023)} Vicuna &69\perc\\
    {\small(2023)} Mistral$_{I.}$ &87\perc\\
    {\small(2023)} Mixtral$_{I.}$ &88\perc\\
    {\small(2024)} Llama-3$_{I.}$ &84\perc\\
    
    \bottomrule    
    \end{tabularx} 
\caption{The level of prompt agreement for each model across three prompts for each time-sensitive question. The agreement is computed as the percentage of times a model gives the same answer to all three prompts. Subscripts ${I.}$ and ${C.}$ stand for \textit{Instruct} and \textit{Chat}, respectively.}
\label{table:promptagreement}
\end{table}

\begin{table*}[h!]
\centering
\small
    \begin{tabular}{lccccccc}
        \toprule
        \multirow{4}{*}{\textbf{Model}} & \multirow{4}{*}{\textbf{\makecell{\#Outdated\\Facts}}} & & \multicolumn{5}{c}{\textbf{Knowledge Editing}} \\
        \cmidrule(rrrr){4-8}
        & & & \multicolumn{2}{c}{\textbf{Modifying Parameters}} & & \multicolumn{2}{c}{\textbf{Preserving Parameters}} \\
        \cmidrule(rr){4-5} \cmidrule(rr){7-8}
        \noalign{}
        & & & {ROME} & {MEMIT} & & {SERAC} & {IKE} \\
        \midrule
        ({\small{2019}}) GPT-2 & 54 & & 12\perc & 25\perc & & 3\perc & 39\perc \\

        ({\small{2021}}) GPT-J & 60 & & 10\perc & 71\perc & & 0\perc & 95\perc\textsuperscript{\textdaggerdbl} \\

        ({\small{2023}}) Llama-2$_{C.}$ & 48 & & 3\perc & 72\perc & & 27\perc & 17\perc \\

        ({\small{2023}}) Mistral$_{I.}$ & 41 & & 0\perc & 0\perc & & --- & 86\perc\textsuperscript{\textdaggerdbl} \\
        \bottomrule
        \end{tabular} 
    \caption{Performance of different methods for aligning the outdated knowledge in 4 LLMs, by \textbf{paraphrase success} \cite{meng2022locating, meng2022mass}. \textbf{\textdaggerdbl} indicates successful alignment of more than 85\% outdated knowledge.}
\label{table:edits_ps}
\end{table*}

\begin{figure*}[ht]
    \centering
    \includegraphics[width=\textwidth]{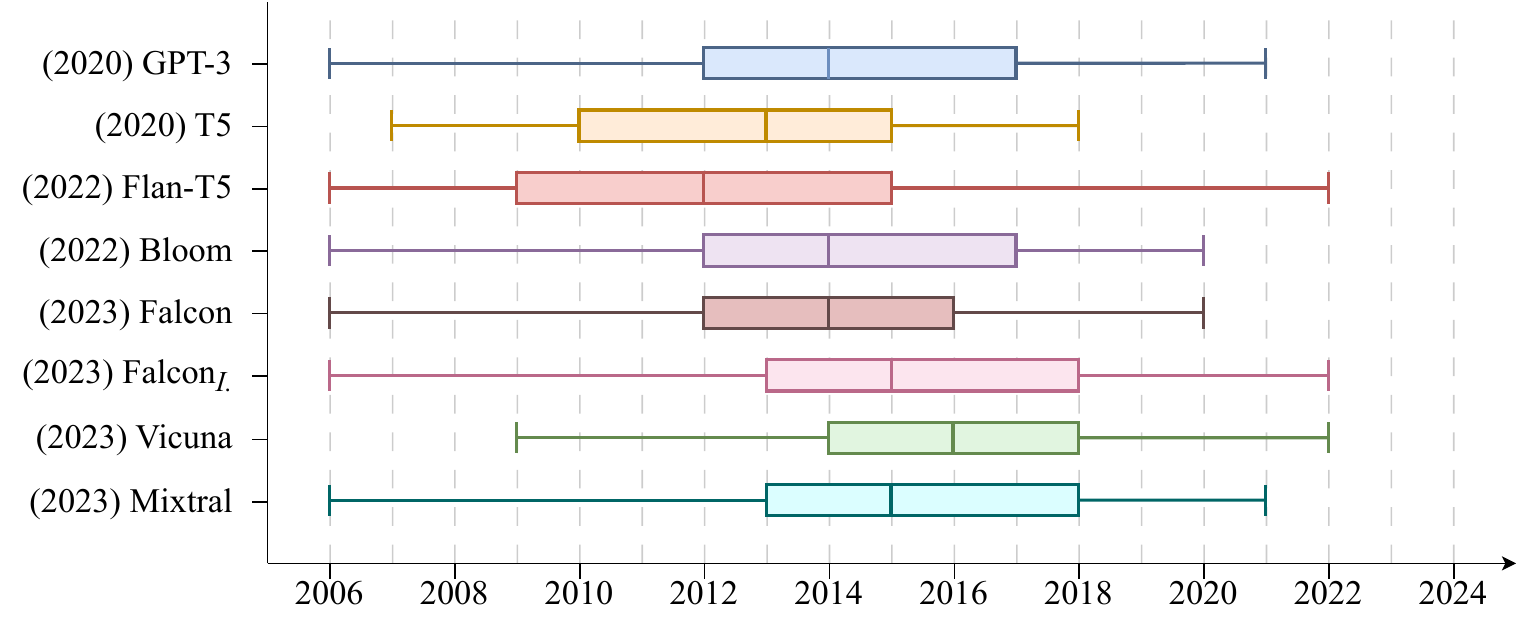}
    \caption{Approximating the temporal period of the data used for (pre-)training the models according to their correct and outdated outputs to our time-sensitive factual questions. The y-axis presents the evaluated LLMs with their release year in parentheses. The box plots present the distribution of the generated responses for each LLM according to their validity interval. Each box plot shows the interquartile range of the responses, with whiskers extending to the minimum and maximum dates.}
    \label{fig:observedyears2}
\end{figure*}

\begin{figure*}[ht]
    \centering
    \includegraphics[width=\textwidth]{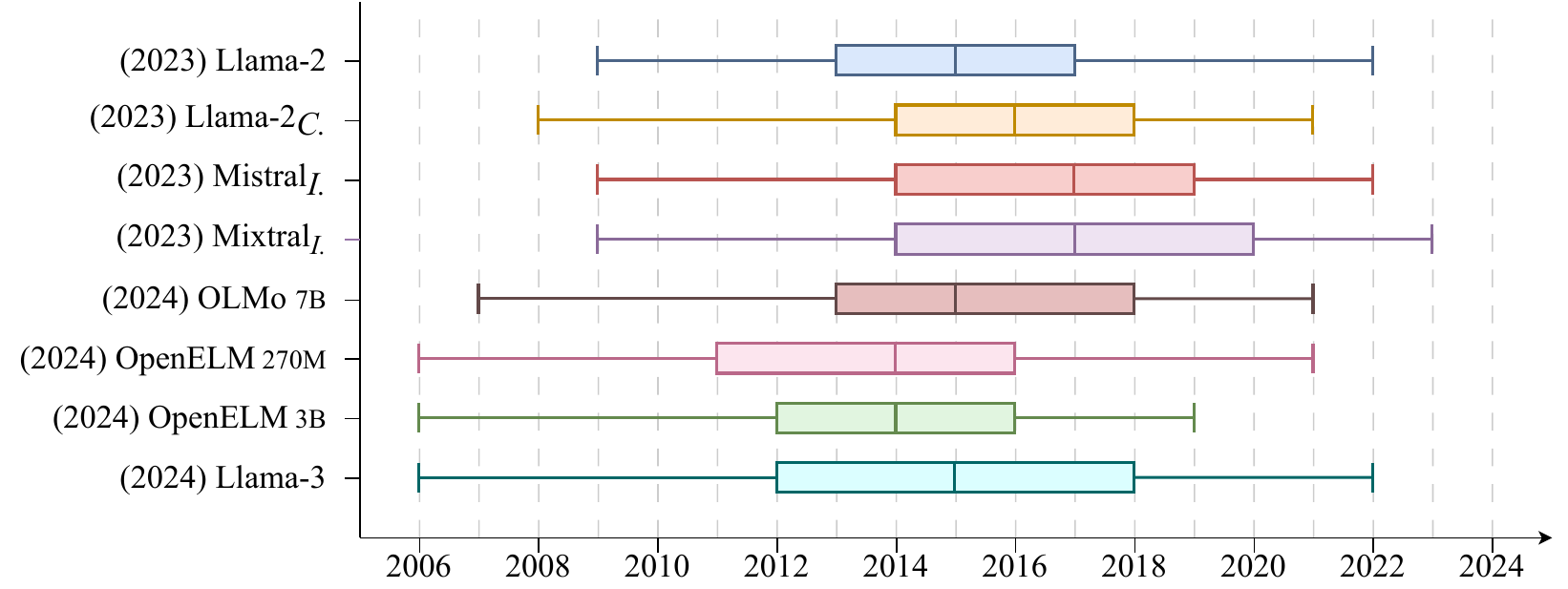}
    \caption{Approximating the temporal period of the data used for (pre-)training the models according to their correct and outdated outputs to our time-sensitive factual questions. The y-axis presents the evaluated LLMs with their release year in parentheses. The box plots present the distribution of the generated responses for each LLM according to their validity interval. Each box plot shows the interquartile range of the responses, with whiskers extending to the minimum and maximum dates.}
    \label{fig:observedyears3}
\end{figure*}

\clearpage

\subsection{Editing Methods Implementation Details}
\label{impdet}

We present more details regarding the editing methods we experiment with:

\begin{itemize}
    \item \textbf{ROME} \cite{meng2022locating} locates the corresponding parameters for each factual knowledge in the feed-forward layers of the model via causal mediation analysis \cite{geva-etal-2021-transformer}. It then inserts a new key-value association (representing the edited knowledge) in the original parameters by formulating it as a least squares problem with a linear equality constraint.
    \item \textbf{MEMIT} \cite{meng2022mass} expands ROME for applying multiple edits at once. While ROME can perform one edit by operating on one layer at a time, MEMIT applies several edits by operating on several layers in one intervention.
    \item \textbf{SERAC} \cite{pmlr-v162-mitchell22a} uses an external memory to store the new facts, and a classifier to measure the similarity of the question prompt with the stored facts in the memory. If there is no match between the question prompt and the facts in the memory, the primary LLM is selected to generate the final output. In cases of a match between the question prompt and a new fact in the memory, a secondary model (a smaller language model) generates the response grounded on the matching new fact. 
    \item \textbf{IKE} \cite{Zheng2023CanWE} is based on in-context learning. To answer the question \textit{q\textsuperscript{*}} by the new (up-to-date) attribute value \textit{y\textsuperscript{*}}, this method constructs a prompt consisting of the question, the corresponding up-to-date fact \textit{f\textsuperscript{*}}, and a context segment (\textit{\textit{q\textsuperscript{*}}, \textit{f\textsuperscript{*}}, C}). The context consists of \textit{k} triplets (\textit{C =} \{\textit{c$_1$,...,c$_k$}\}) of facts, corresponding questions, and values \textit{c$_i$} = (\textit{f$_i$}, \textit{q$_i$}, \textit{y$_i$}). The triplets are retrieved based on the cosine similarity with (\textit{q\textsuperscript{*}}, \textit{f\textsuperscript{*}}, \textit{y\textsuperscript{*}}) from a pre-defined pool and serve as examples for using the information in \textit{f$_i$} to answer \textit{q$_i$}. The prompt (\textit{\textit{q\textsuperscript{*}}, \textit{f\textsuperscript{*}}, C}) is then presented to the model to output an answer. Note that this technique does not represent a realistic scenario, since it requires the relevant and up-to-date fact \textit{f}\textsuperscript{*} for each question to be deterministically provided to the model. 
\end{itemize}

The experiments were applied to Huggingface gpt2-xl, EleutherAI/gpt-j-6b, meta-llama/Llama-2-7b-chat-hf, and mistralai/Mistral-7B-Instruct-v0.1. Regarding the knowledge editing methods, we followed EasyEdit framework \cite{wang2023easyedit} for ROME, MEMIT, SERAC, and IKE and utilized the default model-specific configuration of the hyper-parameters. For Llama-2$_{C.}$ and Mistral$_{I.}$, we considered the configurations for the non-chat and non-instruct versions, respectively. Training SERAC for GPT-2, GPT-J, and Llama-2$_{C.}$ required one NVIDIA A100 with 80 GiB. Model inference was performed on two NVIDIA GeForce RTX 3090 with 24.5 GiB each, except for Mixtral which required three NVIDIA A100 with 80 GiB. Regarding ROME, the edits are applied sequentially and the original weights of the models are not reverted after each edit. That is, as a knowledge repository, the models must be able to retain all the edits. Regarding MEMIT, all the changes are applied in one intervention. Regarding SERAC, the memory consists of the new facts to be edited for each model.  

\end{document}